%% file: main.tex
\documentclass{ecai} 

\usepackage{latexsym}
\usepackage{amssymb}
\usepackage{amsmath}
\usepackage{amsthm}
\usepackage{booktabs}
\usepackage{enumitem}
\usepackage{graphicx}
\usepackage{color}
\usepackage{ragged2e}
\usepackage{multirow}
\usepackage{xcolor}
\usepackage{arydshln}
\usepackage{booktabs}
\definecolor{linkblue}{RGB}{0,90,181}
\usepackage[colorlinks=true, linkcolor=linkblue, urlcolor=linkblue, citecolor=linkblue]{hyperref}

\usepackage[ruled,vlined]{algorithm2e}
\usepackage{tcolorbox}
\newcounter{colorboxcounter}
\newcounter{picturecounter}
\usepackage{acronym}
\acrodef{llm}[LLM]{Large Language Model}
\acrodef{rag}[RAG]{Retrieval-Augmented Generation}
\acrodef{htc}[HTC]{Hierarchical Text Classification}
\acrodef{rag}[RAG]{Retrieval-Augmented Generation}
\acrodef{dag}[DAG]{Directed Acyclic Graph}
\acrodef{kg}[KGs]{Knowledge Graph}

\newcommand{\BibTeX}{B\kern-.05em{\sc i\kern-.025em b}\kern-.08em\TeX}

\begin{document}


\begin{frontmatter}


\paperid{1335} 


\title{{KG-HTC}: Integrating Knowledge Graphs into LLMs for Effective Zero-shot Hierarchical Text Classification}


\author[A]{\fnms{Qianbo}~\snm{Zang} \thanks{Corresponding Author. Email: \href{mailto:qianbo.zang@uni.lu}{qianbo.zang@uni.lu}}}
\author[B]{\fnms{Christophe}~\snm{Zgrzendek}} 
\author[A]{\fnms{Igor}~\snm{Tchappi}}
\author[A]{\fnms{Afshin}~\snm{Khadangi}}
\author[C]{\fnms{Johannes}~\snm{Sedlmeir}}

\address[A]{Interdisciplinary Centre for Security, Reliability and Trust (SnT), Université du Luxembourg}
\address[B]{Enovos Luxembourg S.A.}
\address[C]{Universität Münster}


\begin{abstract}
Hierarchical Text Classification (HTC) involves assigning documents to labels organized within a taxonomy. Most previous research on HTC has focused on supervised methods. However, in real-world scenarios, employing supervised HTC can be challenging due to a lack of annotated data. Moreover, HTC often faces issues with large label spaces and long-tail distributions.
In this work, we present \textbf{K}nowledge \textbf{G}raphs for zero-shot \textbf{H}ierarchical \textbf{T}ext \textbf{C}lassification (KG-HTC), which aims to address these challenges of HTC in applications by integrating knowledge graphs with Large Language Models (LLMs) to provide structured semantic context during classification. 
Our method retrieves relevant subgraphs from knowledge graphs related to the input text using a Retrieval-Augmented Generation (RAG) approach. 
Our KG-HTC can enhance LLMs to understand label semantics at various hierarchy levels. 
We evaluate KG-HTC on three open-source HTC datasets: WoS, DBpedia, and Amazon. Our experimental results show that KG-HTC significantly outperforms three baselines in the strict zero-shot setting, particularly achieving substantial improvements at deeper levels of the hierarchy.
This evaluation demonstrates the effectiveness of incorporating structured knowledge into LLMs to address HTC's challenges in large label spaces and long-tailed label distributions. Our code is available at~\href{https://github.com/QianboZang/KG-HTC}{https://github.com/QianboZang/KG-HTC}.

\end{abstract}

\end{frontmatter}


\input{chapters/introduction}


\input{chapters/method}


\input{chapters/experiment}


\input{chapters/result}


\input{chapters/related_work}


\input{chapters/conclusion}



\begin{ack}
This research was funded in part by the Luxembourg National Research Fund (FNR), grant reference 14783405. 
The research was carried out as part of a partnership with Enovos Luxembourg S.A.
For the purpose of open access, and in fulfillment of the obligations arising from the grant agreement, the author has applied a Creative Commons Attribution 4.0 International (CC BY 4.0) license to any Author Accepted Manuscript version arising from this submission.
\end{ack}



\bibliography{reference}

\clearpage


\end{document}

%% file: chapters/introduction.tex
\section{Introduction}
\label{sec:1}

\begin{figure}[t]
    \centering
    \refstepcounter{picturecounter}
    \label{fig:htc}
    \includegraphics[width=1\linewidth]{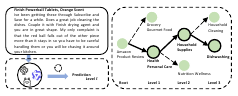}
    \caption{An example of \ac{htc} from the Amazon Product Review dataset.}
    \vspace{20pt}
\end{figure}

Text classification is a fundamental task in natural language processing that focuses on assigning one or more predefined categories to a given piece of text. 
A specific and increasingly important extension of text classification is \ac{htc}, where the labels are not simply flat but are organized within a multi-level taxonomy. 
\ac{htc} aims to categorize textual data into multi-level label systems with parent-child relationships, where labels at different levels are organized as a hierarchical taxonomy. 
Figure~\ref{fig:htc} features an example from the Amazon Product Review dataset. In this figure, the root of the hierarchy is the name of the dataset, and a review needs to be classified to labels at multiple levels. The example input text can be sequentially classified as health personal care, household supplier, and dishwashing. 
\ac{htc} has been successfully applied in different domains, including e-commerce product categorization \citep{chatterjee-etal-2021-joint, patel-2022-modeling}, hierarchical topic modeling of e-participation platforms \citep{simonofski-2021-supporting}, and fine-grained literature management \citep{chen-etal-2021-hierarchy, kowsari-2017-hdl}. 

However, in real-world applications, \ac{htc} often faces one or multiple out of the following three significant challenges. 
First, there may be a shortage of annotated data, particularly as the cost of manually labeling custom data at multiple hierarchical levels is prohibitively high \citep{chen-2022-clean}. 
This problem becomes even more severe in dynamic environments such as retail systems, where taxonomies evolve with new product lines. 
Second, practical taxonomies universally exhibit large-scale label spaces \citep{zhang-2024-teleclass}. 
For instance, the hierarchy of the Amazon Product Review dataset contains over 500~leaf categories.\footnote{\href{https://www.kaggle.com/datasets/kashnitsky/hierarchical-text-classification}{kaggle.com/datasets/kashnitsky/hierarchical-text-classification}}   
Third, real-world datasets in \ac{htc} can exhibit highly imbalanced long-tail distributions, i.e., a small number of frequent categories dominates the dataset while most classes remain underrepresented. 
For instance, our statistical analysis reveals that the most frequent 15\,\% of categories account for 80\,\% of total instances in the third level of the Amazon Product Review dataset. In contrast, the bottom 50\,\% of categories collectively represent merely 6\,\% of the data volume.

\begin{figure*}[t]
    \centering
    \refstepcounter{picturecounter}
    \label{fig:pipeline}
    \includegraphics[width=1\linewidth]{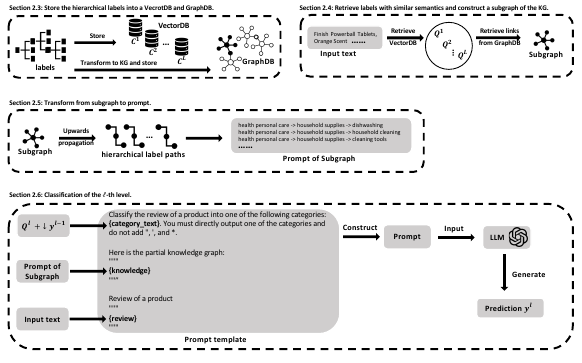}
    \caption{The overview pipeline of KG-\ac{htc}.}
    \vspace{25pt}
\end{figure*}

For these three reasons, and contrary to flat text classification, standard supervised approaches are not well-suited to \ac{htc} in many industry settings \citep{halder-2020-task, huang-2019-hierarchical, kowsari-2017-hdl, meng-2019-weakly}. 
As a consequence, researchers have increasingly turned to zero-shot learning methods to approach \ac{htc}~\citep{bongiovanni-2023-zero, halder-2020-task, paletto-etal-2024-label}. We can group recent studies into three main approaches. 
The first approach leverages the in-context learning of pre-trained LLMs through carefully designed prompts to accomplish accurate classification. 
\citet{halder-2020-task} transforms any text classification task into a universal binary classification problem, where LLMs assign a binary value (true or false) for each label in the label space. However, due to \ac{htc}'s extensive label space, this method requires multiple iterations to complete a single classification. 
The second approach utilizes embedding models converted from pre-trained \ac{llm} to calculate the distance between input text and labels for classification. 
\citet{bongiovanni-2023-zero} proposed Z-STC to propagate similarity scores up the hierarchy and leverage this propagated information to optimize classification for upper-level labels. 
The third method combines embedding models with LLM-based classification. 
\citet{paletto-etal-2024-label} introduced HiLA, where \acp{llm} generate new label layers inserted into the bottom of the current taxonomy. Then, Paletto follows Z-STC for classification. 
However, \citet{bongiovanni-2023-zero}'s and \citet{paletto-etal-2024-label}'s methods exhibit low classification performance for deeper-level labels.

In this paper, we propose KG-\ac{htc}, a new method to enhance \ac{htc} by integrating \ac{kg} into \ac{llm} for \ac{htc}. To this end, we will focus on the \textbf{strict zero-shot learning} setting. 
Since \ac{dag} can represent the hierarchical structure of labels in \ac{htc}, the combination of \ac{llm} and knowledge graphs is especially promising for this purpose. 
We represent the taxonomy as a DAG-based knowledge graph and compute the cosine similarity between the text and the embeddings of labels at each level. 
By applying preset thresholds, candidate labels that are highly semantically relevant to the input text are chosen at every hierarchical level. 
Then, leveraging these candidate labels, the system dynamically retrieves the most pertinent subgraph from the complete label knowledge graph corresponding to the given text. 
For the retrieved subgraph, an upwards propagation algorithm is employed to systematically enumerate all possible hierarchical paths from the leaf nodes to the root, with each path representing a complete reversed hierarchical label sequence. 
These structured sequences are subsequently concatenated into a prompt, which is fed into an LLM to perform the zero-shot classification task. 

We evaluate our approach using three public datasets and achieve new state-of-the-art results for all of them.
Without relying on any annotated data, the KG-HTC method significantly enhances the model's capability to overcome the issue of long-tail and sparse labels. We find that compared to the approaches of \citet{bongiovanni-2023-zero} and \citet{paletto-etal-2024-label}, our KG-\ac{htc} exhibits significantly smaller performance degradation as the label hierarchy deepens. As such, the contributions of this paper are threefold:
\begin{enumerate}
    \item We present a novel approach -- KG-HTC -- to integrate \ac{kg} into \ac{llm} for \ac{htc} under a strict zero-shot setting.
    \item We implement a novel pipeline that semantically retrieves relevant subgraphs from the label taxonomy based on cosine similarity with the input text and transforms it to a structured prompt.
    \item We conduct an in-depth assessment of KG-HTC, comparing it with the previous state-of-the-art on three open-source \ac{htc} datasets.
\end{enumerate}

The remainder of this paper is structured as follows: We introduce KG-HTC in Section~\ref{sec:2}. In Section~\ref{sec:3}, we describe our experimental settings, followed by an analysis of our experimental results in Section~\ref{sec:4}. 
We survey related work in Section~\ref{sec:5}. 
Finally, we conclude our research in Section~\ref{sec:6}.

%% file: chapters/method.tex
\section{Method}
\label{sec:2}

\subsection{Problem Identification}
\label{sec:2.1}

Zero-shot text classification via \ac{llm} is formally framed as a generative classification task. 
Given an input text sequence \( x = (x_1, x_2, \dots, x_n) \), where \( n \) denotes text length, an LLM generates an output text \( y \sim \mathcal{LLM}(x) \) using a Top-p sampling strategy. 
In classification contexts, the set of labels can be defined as \( C \).
The generated text \( y \) can then be mapped to a predicted label \(\hat{c} \in C\).

\input{tables/symbols}

\begin{figure}[t]
    \centering
    \refstepcounter{picturecounter}
    \label{fig:visual_graph}
    \includegraphics[width=1\linewidth]{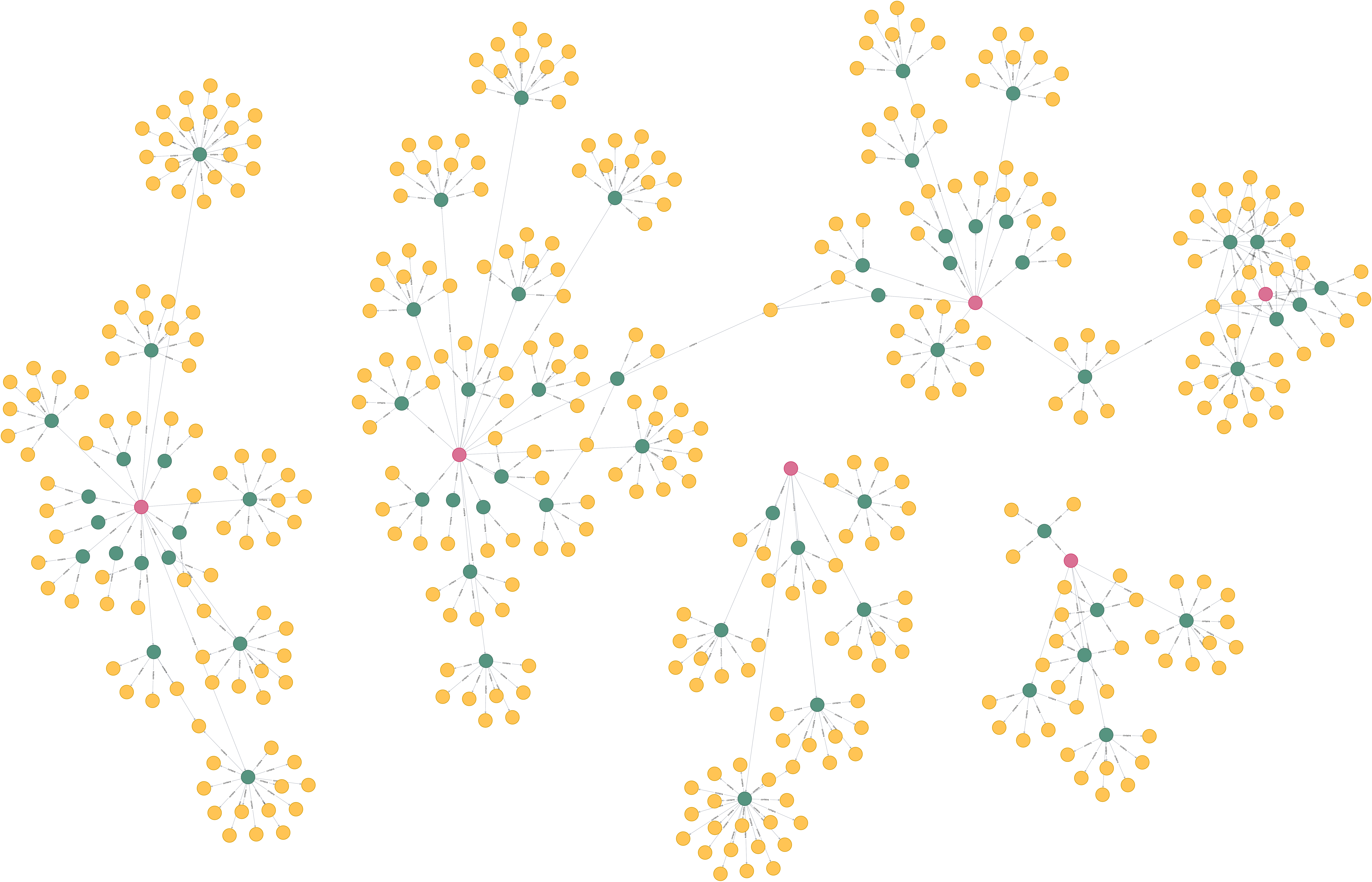}
    \caption{Visualization of the knowledge graph (tree) constructed from the multi-level taxonomy in the Amazon Product Review dataset. The red nodes represent labels in the first hierarchical level. The green nodes denote sub-categories (second level) interconnected through parent-child relationships. And the yellow nodes correspond to finally fine-grained leaf categories in the third level.}
    \vspace{25pt}
\end{figure}

Our mathematical framework follows the formulation established by \citet{paletto-etal-2024-label}, with key symbols  summarized in Table~\ref{tab:symbols}. 
In hierarchical text classification, all labels \( C \) in the label space are organized into a hierarchical taxonomy \( C = (C^1, C^2, \dots, C^L) \), where \( L \) is the maximum depth of the taxonomy, i.e., the number of levels.
To formally capture the dependencies of the labels, we adopt the upward and downward arrow notation employed by~\citep{paletto-etal-2024-label}, where \( \uparrow c_i^l \) represents the parent node of \( c_i^l \) in the hierarchy and \( \downarrow c_i^l \) represents the child nodes of \( c_i^l \) in the hierarchy.
For all $l>1$, each label \( c_i^l \in C^l \) must satisfy the structural constraint \( \uparrow c_i^l \in C^{l-1}\).
The task of hierarchical text classification requires an \ac{llm} to iteratively generate outputs \( (y^1, y^2, \dots, y^L) \) corresponding to predicted labels \( (\hat{c}^1, \hat{c}^2, \dots, \hat{c}^L) \) within a hierarchical taxonomy \( \hat{c}^l \in C^l\). 

\subsection{System architecture}
\label{sec:2.2}

Figure~\ref{fig:pipeline} illustrates the full process of our KG-\ac{htc}. 
First, we store all labels into a graph database and a vector database, respectively. 

Given an input text, we secondly retrieve labels of top candidates \( Q^l \) at each level \( l \) from the vector database and retrieve a valid subgraph from the graph database by checking parent-child relationships between candidates from adjacent levels. 
Then, we convert a set of paths in a retrieved subgraph into a structured prompt and concatenate the structured context with a classification prompt. 
Finally, we leverage In-context Learning for zero-shot text classification. 

\input{algorithms/subgraph}

\subsection{Storage of hierarchical labels}
\label{sec:2.3}
In the first step, we store all labels into a graph database and a vector database respectively. 
In hierarchical text classification, the taxonomies of labels can be conceptualized as DAG knowledge graphs, where multi-level labels are interconnected through hierarchical affiliation relationships. 
By explicitly defining the relational pathways between labels in each tier, \ac{llm} can develop a structural understanding of individual labels and their conceptual boundaries within the hierarchy. 
For instance, the input from Figure~\ref{fig:visual_graph} can be classified as dishwashing or cleansers.
However, cleansers is the child node of bath body in the taxonomy of the Amazon Product Review dataset, which means the conceptual boundary of cleansers is within human cleansers.
In this case, the \ac{llm} can classify the input to dishwashing as the correct answer.
This graph-based knowledge representation equips \ac{llm}s with dual advantages: it not only establishes explicit semantic navigation pathways for text processing tasks, but also creates topological constraints that may substantially enhance classification accuracy in zero-shot scenarios through improved semantic disambiguation~\citep{liu-2023-enhancing}.

\subsection{Subgraph Retrieval}
Empirical studies have demonstrated that the \ac{rag} framework exhibits significant advantages in open question-answering tasks \citep{gao-2023-retrieval, karpukhin-etal-2020-dense, lewis-2020-retrieval}. 
A defining challenge in \ac{htc} stems from the scenarios of large label space, where the label taxonomy often spans massive categories.
By dynamically retrieving relevant documents from vector databases through a similarity check with in-context learning, \ac{rag} effectively enhances the factual accuracy of generated responses.

The research of \citet{li-2024-long} shows that \ac{llm} exhibit limitations in both long-context processing and classification tasks of large label space. 
Directly encoding the full knowledge graph into LLMs may therefore suffer from performance degradation due to information overload or attention dilution. 
To mitigate this, we propose an \ac{rag} enhanced framework that dynamically retrieves semantically relevant subgraph (sub-tree) components from the whole knowledge graph based on input text. 
These retrieved subgraphs are subsequently structured as contextualized prompts, enabling the classifier to prioritize critical hierarchical dependencies while suppressing irrelevant noise. We will thoroughly introduce this in Section~\ref{sec:prompt}.

Specifically, for each hierarchical level \( l \), we compute the cosine similarity distance between the input text embedding and all labels at level \( l \): 
\begin{equation}  
\label{eq:distance}  
d_{C}(x, \ c_i^l) = 1 - S_{C}\left( \Psi(x), \, \Psi(c_i^l) \right),  
\end{equation}  
where \( \Psi: X \to \mathbb{R}^d \) denotes the embedding function, and \( S_{C}(\cdot, \ \cdot) \) represents the cosine similarity operator. We then retrieve labels of top candidates at each level \( l \):
\begin{equation}  
\label{eq:query}  
Q^l = \{ c_i^l \, | \, d_{C}(x, \ c_i^l) \leq \tau_l \},  
\end{equation} 
where \( \tau_l \) is a similarity threshold.  

To ensure hierarchical consistency, we validate cross-level dependencies by checking parent-child relationships between candidates from adjacent levels.
A candidate label \( c_i^{l} \in Q^l \) must satisfy the condition that its parent label \( \uparrow c_i^{l} \) belongs to \( Q^{l-1} \). Algorithm~\ref{alg:subgraph} shows the complete process for retrieving a subgraph based on input text \(x\) as the query.

\subsection{Transformation from Subgraph to Prompt}
\label{sec:prompt}
For an input text \( x \), we first retrieve a subgraph \( G \) through Algorithm~\ref{alg:subgraph}. 
To effectively inject the KG into the LLM, we adopt a strategy that converts one graph structure into a set of paths through parent-child edge connections \citep{zhang-2024-knowgpt}.
Specifically, each subgraph is serialized as a set of hierarchical paths from root to leaf nodes. 
This chain-based representation ensures structural consistency while maintaining compatibility with sequential input formats of \ac{llm}. 
For \ac{htc} tasks, the length of paths equals the taxonomy depth \( L \).

\input{algorithms/to_prompt}
\ac{llm} cannot directly understand graph information. We need to transfer the subgraph to a prompt.
A feasible method is to input the paths inside graphs as a prompt for graph
understanding and reasoning~\cite{luo-2024-graphinstruct}.
Valid hierarchical label paths in the subgraph are preserved as contextual prompts with the upwards propagation. 
The upwards propagation is implemented as a loop traversal algorithm that systematically explores all possible paths from the terminal leaf nodes to the root node in a hierarchical structure. 
Starting from any initial node \(\forall c_i^L \in C^L\) in the final hierarchical level, the Algorithm~\ref{alg:kg_chains} can progressively visit father nodes until reaching root nodes, then backtracks to explore alternative branches. 
This exhaustive traversal guarantees the complete enumeration of all valid path combinations.
After obtaining all path combinations, the sequence of elements in each path \( P_i\) will be reversed to ensure the directional consistency from the first layer to the \(L\)-th layer, with the final output being the complete set of paths \( P= \{ P_1, \ P_2, \ ... \ \} \). 
Each path \(P_i = (p_i^1 \rightarrow p_i^2 \rightarrow ... \rightarrow p_i^L) \) represents a connected node sequence, where \( p_i^l \in G \) and \(p_i^l \in C^l\).

We can convert a set of paths in a retrieved subgraph into a structured prompt using \( \rightarrow \) to connect two nodes in adjacent levels.
This structured context enables the LLM to recognize hierarchical constraints during classification.
An illustrative prompt example of such multi-level label paths for the Amazon Product Review dataset is provided in Prompt~\ref{box:kg}.

\input{algorithms/classification}
\label{sec:classification}
\subsection{Classification of Each Level}

Our approach leverages In-context Learning and prompt engineering by concatenating the structured context extracted from the retrieved subgraph with a classification prompt template. And then, the  concatenated prompt will be fed into the LLM for inference. 
Prompt~\ref{box:sys_prompt} is an example of the classification template from the Amazon Product Review dataset.

For \ac{htc}, we follow a layer-wise classification strategy. The model begins by predicting the label at the first level, then proceeds to classify labels at deeper levels sequentially, until the final level is reached. 
One major challenge in HTC is the large number of candidate labels. 
Providing all possible labels in a single prompt often leads to performance degradation \cite{li-2024-long}. 
However, since the number of first-level labels is usually small (no more than 10 among our evaluation datasets), we can directly include all first-level labels into the prompt. 
This strategy simplifies inference and enhances classification performance.

For classification beyond the first level, the model uses the prediction from the previous level as a constraint for the current level. 
For instance, the predicted label of level \(l\) is \(y^l\).
The candidate label set at the next level \(l + 1\) mainly consists of the subcategories \(\downarrow y^l\) of the previous prediction \(y^l\), along with additional labels \( Q^{l+1} \) retrieved using Equation~\ref{eq:query}.
This equation helps mitigate the impact of potential errors in the previous level’s prediction and improves the overall robustness and accuracy of the classification process. 
Algorithm~\ref{alg:htc} illustrates the full process of our KG-\ac{htc}.

\input{boxes/partial_graph}
\input{boxes/classification}

%% file: tables/symbols.tex
\begin{table}[t]
\renewcommand{\arraystretch}{1.4}
\caption{List of symbols.}
\vspace{10pt}
\centering
\begin{tabular}{ll} 
\toprule
\textbf{Symbol} & \textbf{Explanation} \\
\toprule
\( l \)                     & A hierarchy level of a label taxonomy, \( l \in (1, \ ... \ , L)\) \\
\( C^l \)                   & Set of all labels at level \( l \) \\
\( c_i^l \)                 & A label at level \( l \) \\
\( \uparrow c_i^l \)        & The parent label of \( c_i^l \) \\
\( \downarrow c_i^l \)        & The set of child labels of \( c_i^l \) \\
\( \Psi(x) \)               & A vector embedding of a text \( x \) \\
\( S_C(A, \ B) \) & The cosine similarity of two embeddings \( A \) and \( B \) \\
\( Q^l \)         & All queried labels by \( S_C \) with a threshold \( \tau_l \) at level \( l \) \\
\bottomrule
\end{tabular}
\label{tab:symbols}
\vspace{5pt}
\end{table}

%% file: algorithms/subgraph.tex
\begin{algorithm}[t]
\caption{Subgraph Retrieval}
\label{alg:subgraph}
\SetAlgoLined 
\KwIn{Input text $x$, labels \( \{ C^1, C^2, \ldots, C^L \} \)}
\KwOut{Retrieved sub-graph $G$}
\BlankLine
Initialize an empty graph  structure $G \leftarrow \emptyset$\;
Compute the embedding for the input text $\Psi(x)$\;
\For{ $l \leftarrow 1$ to $L$ }{
    Compute $\Psi(c_i^l)$ for all labels at level $l$\;
    Calculate $d_{C}$ via Equation~\ref{eq:distance}\;
    Retrieve candidates $Q^l$ via Equation~\ref{eq:query}\;
}
\For{ $l \leftarrow 2$ to $L$ }{ 
    \For{each \(c_i^l \in Q^l\) }{
        \If{\( \uparrow c_i^l \in Q^{l-1}\) }{
            Add nodes \( \uparrow c_i^l \) and $c_i^l$ to $G$\;
            Add edge $(\uparrow c_i^l, \ c_i^l)$ to $G$\;
        }
    }
}
Delete repeated \((\uparrow c_i^l, \ c_i^l) \) in \(G\) \;
\Return $G$\;
\end{algorithm}

%% file: algorithms/to_prompt.tex
\begin{algorithm}[t]
\caption{From Subgraph to Hierarchical Label Paths}
\label{alg:kg_chains}
\SetAlgoLined
\KwIn{Input text $x$, subgraph $G$ retrieved via Algorithm~\ref{alg:subgraph}}
\KwOut{Set of hierarchical paths $\{ P_1, P_2, ... \}$}

\BlankLine

Initialize path set $P \gets \emptyset$;

\For{each leaf node $\forall c_i^L \in C^L$}{
    Initialize a stack structure $S \gets \{[c_i^L]\}$;
    
    \For{$l \leftarrow L$ to $1$}{
        Push the parent node $\uparrow c_i^l \in C^{l-1}$ of $c_i^L$ to the \(S\);
    }
    
    Add S to P;
}
\Return $P$;
\end{algorithm}

%% file: algorithms/classification.tex
\begin{algorithm}[t]
\caption{KG-\ac{htc}}
\label{alg:htc}
\SetAlgoLined 
\KwIn{Input text \( x \), prompt template $P$}
\KwOut{Predicted hierarchical labels $\mathbf{y} = (y^1, y^2, ..., y^L)$}
\BlankLine
Initialize the set of predicted labels $y \gets \emptyset$\;
Retrieve subgraph $G$ via Algorithm~\ref{alg:subgraph}\;
Transfrom \(G\) to hierarchical label paths \(P\)\;

\For{level $l \gets 1$ to $L$}{
    \eIf{$l \gets 1$}
    {
        $prompt \gets$ $\textit{concat}(x, \ P, \ C^1)$ via Prompt~\ref{box:sys_prompt}\;
    }{
        Retrieve candidates $Q^l$ via Equation~\ref{eq:query}\;
        $prompt \gets$ $\textit{concat}(x, \ P, \ Q^l  \cup \downarrow y^{l-1})$ via Prompt~\ref{box:sys_prompt}\;
    }
    \(y^l \gets \mathcal{LLM} (Prompt)\)\;
    Add \(y^l\) into \(y\) \;
}
\Return $\mathbf{y} = (y^1, ..., y^L)$
\end{algorithm}
 

%% file: boxes/partial_graph.tex
\begin{figure}[t]
\centering
\begin{tcolorbox}[
    colback=blue!5!white,
    colframe=blue!75!black,    
    title={\textbf{Prompt 1: A Retrieved Subgraph for Amazon Product Review Dataset}},
    boxrule=2pt,
    arc=1mm,
    fontupper=\fontsize{7.5}{10}\selectfont,    
    fonttitle=\fontsize{7}{6.5}\selectfont\bfseries, 
    boxsep=1pt,
]
\refstepcounter{colorboxcounter}
\label{box:kg}
pet supplies -> cats -> cat flaps\\
pet supplies -> dogs -> doors\\
pet supplies -> cats -> collars\\
pet supplies -> cats -> litter housebreaking\\
baby products -> safety -> gates doorways\\
baby products -> safety -> harnesses leashes\\
baby products -> safety -> bathroom safety\\
pet supplies -> cats -> feeding watering supplies\\
baby products -> safety -> cabinet locks straps\\
toys games -> stuffed animals plush -> plush puppets\\
pet supplies -> dogs -> houses\\
baby products -> safety -> sleep positioners\\
baby products -> potty training -> step stools\\
toys games -> stuffed animals plush -> plush backpacks purses\\
pet supplies -> cats -> carriers strollers\\
baby products -> gear -> swings
\end{tcolorbox}
\end{figure}

%% file: boxes/classification.tex
\begin{figure}[t]
\centering
\begin{tcolorbox}[
    colback=blue!5!white,
    colframe=blue!75!black,    
    title={\textbf{Prompt 2: Classification Template for Amazon Product Review Dataset}},
    boxrule=2pt,
    arc=1mm,
    boxsep=1pt,
    fontupper=\fontsize{7.5}{10}\selectfont,    
    fonttitle=\fontsize{7}{6.5}\selectfont\bfseries, 
]
\refstepcounter{colorboxcounter}
\label{box:sys_prompt}
Classify the review of a product into one of the following categories: \textbf{\{category\_text\}}. You must directly output one of the categories and do not add ", ', and *.\\

Here is the partial knowledge graph: \\
"""\\
\textbf{\{knowledge\}}\\
"""
\end{tcolorbox}
\end{figure}

%% file: chapters/experiment.tex
\section{Experiment}
\label{sec:3}

\subsection{Dataset}

\paragraph{Amazon Product Reviews (Amazon) \citep{yury_2020_amazon}. \protect \footnote{\href{https://www.kaggle.com/datasets/kashnitsky/hierarchical-text-classification}{kaggle.com/datasets/kashnitsky/hierarchical-text-classification}}}~This dataset contains reviews of products on the Amazon application. 
Each data item contains a title and a description and needs to be classified according to a three-level hierarchical taxonomy with 6, 64, and 510 label categories, respectively.

\paragraph{Web of Science (WoS) \citep{kowsari-2017-hdl}. \protect \footnote{\href{https://huggingface.co/datasets/HDLTex/web_of_science}{huggingface.co/datasets/HDLTex/web\_of\_science}}}~This dataset is about scientific research literature. 
It includes data from multiple fields such as natural sciences, social sciences, and humanities and arts, and is widely used in academic research, bibliometric analysis, and scientific evaluation. 
Each data item is classified according to a two-level hierarchical taxonomy, encompassing 7 and 134 label categories at each level, respectively.

\paragraph{Dbpedia \citep{auer-2007-dbpedia}. \protect \footnote{\href{https://www.kaggle.com/datasets/danofer/DBpedia-classes}{kaggle.com/datasets/danofer/DBpedia-classes}}}~This dataset is an open knowledge base project built on Wikipedia. 
Scientists extract and transform the vast wealth of information from Wikipedia and present it in a structured and standardized format. 
The data within the DBpedia forms a massive and complex knowledge graph that supports users in cross-domain knowledge exploration. 
Each data item is classified according to a three-level hierarchical taxonomy, encompassing 9, 70, and 219 label categories at each level, respectively.

\input{tables/main_result}

\begin{figure*}[t]
    \centering
    \refstepcounter{picturecounter}
    \label{fig:f1_decay}
    \includegraphics[width=1\linewidth]{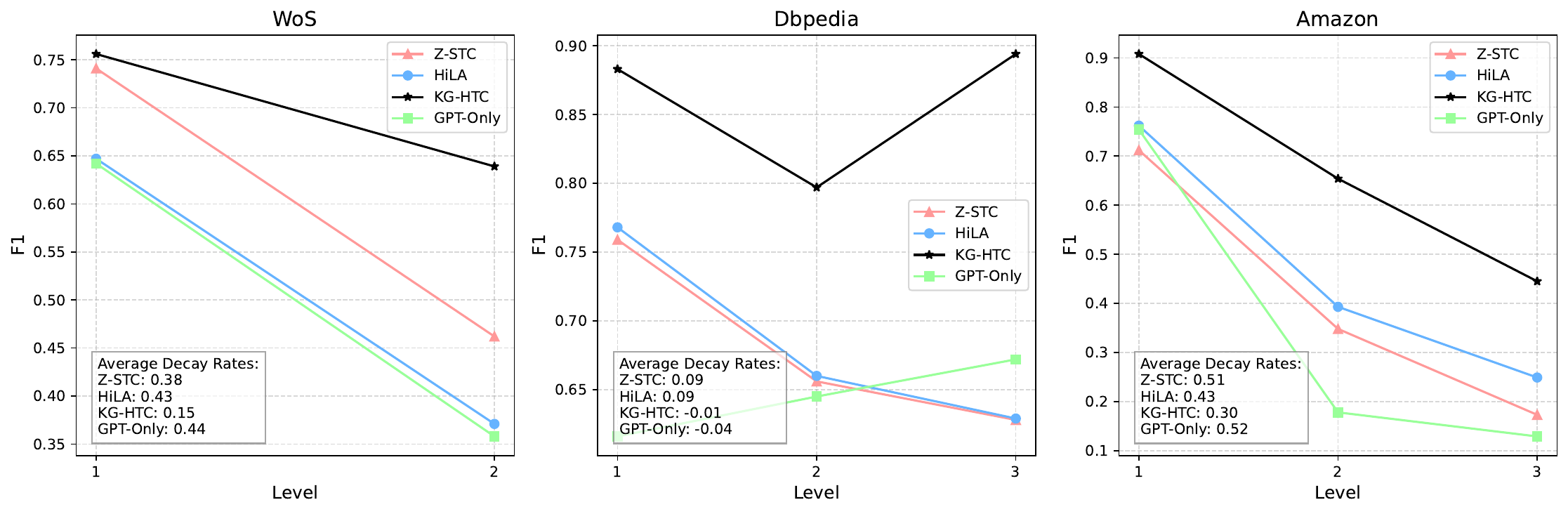}
    \caption{As the taxonomy deepens, KG-HTC exhibits a slower performance degradation on the WoS and Amazon datasets.}
    \vspace{25pt}
\end{figure*}

\subsection{Metric}
 We employ the F1-macro score as the evaluation metric, which is calculated by taking the arithmetic mean of the F1-scores across all classes. The F1-macro ensures that rare classes contribute equally to the final metric. This prevents overoptimistic performance estimates in scenarios where certain classes are underrepresented, making it suitable for imbalanced datasets. Formally, it is defined as:  
\begin{equation}
\text{F}_1\text{-macro} = \frac{1}{C} \sum_{i=1}^{C} {\text{F}_1}_i
\end{equation} 
where \( C \) denotes the total number of classes, and the F1-score for the \( i \)-th class is the harmonic mean of precision and recall:  
\begin{equation}
{\text{F}_1}_i = 2 \times \frac{\text{Precision}_i \times \text{Recall}_i}{\text{Precision}_i + \text{Recall}_i}
\end{equation} 

To evaluate the compounded challenges of large label spaces and long-tailed distributions, we implement a metric called average decay rate via Algorithm~\ref{alg:decay}. Because the label spaces become larger and the datasets become more biased as the levels deepen in the taxonomy.

\begin{equation}
\text{decay}_i = \frac{({\text{F}_1}\text{-macro}\text{ of level}_{i-1}) - ({\text{F}_1}\text{-macro}\text{ of level}_{i})}{({\text{F}_1}\text{-macro}\text{ of level}_{i-1})}
\end{equation}

\begin{equation}
\label{alg:decay}
\text{decay}_{\text{avg}} = \frac{1}{L-1}\sum_{i=2}^{L} \text{decay}_i
\end{equation}

\subsection{Experiment setup}

\paragraph{Baselines.}~We evaluate our method by comparing the results to two baselines. The first baseline is a weak baseline, where \ac{llm} will directly classify each data point per each layer of label. 
Secondly, we will use two previous studies Z-STC and HiLA introduced in Section 2 as strong baselines~\citep{bongiovanni-2023-zero, paletto-etal-2024-label}. These two studies were the previous state-of-the-art in the strict zero-shot setting. 

\paragraph{Experiment details.}~We choose the \texttt{GPT-3.5-turbo} model to align with the research of our strong baselines~\citep{paletto-etal-2024-label}.~\footnote{\href{https://platform.openai.com/docs/models/gpt-3.5-turbo}{platform.openai.com/docs/models/gpt-3.5-turbo}}
We choose the \texttt{text-embedding-ada-002} as the embedding model for \ac{rag} system.~\footnote{\href{https://platform.openai.com/docs/guides/embeddings}{platform.openai.com/docs/guides/embeddings}}
We choose the \texttt{neo4j} as the graph database in our experiments.~\footnote{\href{https://neo4j.com}{neo4j.com}}
We choose the \texttt{ChromaDB} as the vector database in our experiments.~\footnote{\href{https://www.trychroma.com}{trychroma.com}}

In our experiments among all datasets, we set the temperature of the \texttt{GPT-3.5-turbo} model to \(0.4\) and the Top-p parameter to \(0.4\). 
We use a low temperature and a low Top-p value for stable response of \texttt{GPT-3.5-turbo}.
For the \ac{rag} system, we selected similarity threshold values \( \tau_l \) such that there are 10 label candidates retrieved at the second level and 40 at the third level between Dbpedia and Amazon datasets. We selected similarity threshold values \( \tau_l \) such that there are 20 label candidates retrieved at the second level for the WoS dataset. Those hyper-parameters of \ac{rag} were chosen experimentally.

Due to the inherent randomness in the generation of \ac{llm}, the generated output may occasionally fall outside the predefined label space. 
In these cases, we convert the invalid output to a valid label by randomly sampling from the label space, using a fixed random seed of 42 for reproducibility.

%% file: tables/main_result.tex
\begin{table*}[t]
\renewcommand{\arraystretch}{1.5}
\caption{Main results. The evaluation metric is F1-macro. Our KG-HTC provides consistent and significant improvements over both the weak baseline and the strong baselines.}
\vspace{10pt}
\centering
\resizebox{\textwidth}{!}{
\begin{tabular}{c c c c c c c c c c}
\toprule  
\multicolumn{2}{c}{\multirow{2}{*}{\textbf{Model}}}     & \multicolumn{2}{c}{\textbf{WoS}}      & \multicolumn{3}{c}{\textbf{Dbpedia}}     & \multicolumn{3}{c}{\textbf{Amazon}} \\ 
\cmidrule(lr){3-4}
\cmidrule(lr){5-7}
\cmidrule(lr){8-10}
                 &               & \textbf{Level 1}  & \textbf{Level 2}  & \textbf{Level 1}  & \textbf{Level 2} & \textbf{Level 3} & \textbf{Level 1} & \textbf{Level 2} & \textbf{Level 3} \\ 
\toprule
\textbf{Weak baselines}   & GPT-3.5-turbo &    0.642     &    0.358     &   0.616      &     0.645    &    0.672     &      0.754   &    0.178     &    0.129     \\ 
\hdashline
\multirow{2}{*}{\textbf{Strong baselines}} & Z-STC~\citep{bongiovanni-2023-zero}         & 0.741   & 0.462   & 0.759   & 0.656   & 0.628   & 0.712   & 0.348   & 0.173   \\ 
\cdashline{2-10}
                 & HiLA~\citep{paletto-etal-2024-label}          & 0.647   & 0.371   & 0.768   & 0.660   & 0.629   & 0.762   & 0.393   & 0.249   \\ 
\hdashline
\textbf{Ours}             & \textbf{KG-HTC}        &  \textbf{ 0.756}      &    \textbf{0.630}     &     \textbf{0.883}    &      \textbf{0.796}   &   \textbf{ 0.894 }    &    \textbf{0.908 }    &   \textbf{0.654 }     &   \textbf{0.445 }     \\ 
\toprule
\end{tabular}
}
\label{tab:main_results}
\vspace{5pt}
\end{table*}

%% file: chapters/result.tex
\section{Results}
\label{sec:4}

\input{tables/error}

\subsection{Main Results}

The experimental results in Table~\ref{tab:main_results} demonstrate that our KG-HTC provides consistent and significant improvements over both the weak baseline and the strong baselines. 

Compared to using \texttt{GPT-3.5-turbo} alone (our weak baseline) for zero-shot classification, KG-HTC demonstrates remarkable performance improvements. 
Experimental results show that the average performance improvement is 27.1\% for the first level classification, while the second and third levels achieve enhancements of 123.1\% and 139.0\% respectively. 
These results validate that integrating knowledge graphs into \ac{llm} significantly enhances the performance of \ac{htc}. 
As the classification level increases, KG-HTC demonstrates progressively larger improvements, particularly in handling high-level abstract information. This indicates that KG-HTC can effectively address the challenges associated with large label space and long-tailed distributions in hierarchical classification.

As we illustrate in Figure~\ref{fig:f1_decay}, our KG-HTC indicates the lowest performance degradation on the WoS and Amazon datasets as the taxonomy deepens, and the performance gap among the whole three baselines significantly widens. 
These findings further demonstrate that our approach effectively addresses the challenges posed by large label space and long-tailed distributions in hierarchical classification.

\subsection{Error Analysis}

Hit@K measures the fraction of queries for which the correct document appears in the top K results returned by the retriever. 
We described the details of our RAG system in Section~\ref{sec:3}.
For Dbpedia and Amazon, we use Hit@10 for the classification of the second level and Hit@40 at the third level. 
We further use Hit@20 for the classification of the second level of WoS.  
From the results in Table~\ref{tab:error}, we observe that the Hit@K of the \ac{rag} system decreases at the second and third levels across all misclassified samples. 
This decline suggests that one key reason for performance degradation is the RAG system's inability to retrieve the correct subgraph during inference.
When the retrieved knowledge is not closely aligned with the input text or the classification task, the model lacks sufficient contextual support for accurate reasoning. 
As a result, it becomes more likely that incorrect labels are chosen. Inaccurate or irrelevant subgraphs may also introduce noise or confusion, further reducing the model’s ability to assign the correct labels.
This observation also implies that the performance of KG-HTC can continue to improve as information retrieval techniques evolve, especially in terms of retrieval precision and relevance.

\subsection{Ablation study}

\input{tables/ablation_rag}

We sampled a subset with 5000 data samples for each evaluation dataset with a random seed of 42 in the following experiments.

\paragraph{Effectiveness of the Subgraph Retrieval.} Our empirical results indicate that involving knowledge graphs via \ac{rag} can improve the performance of \ac{llm} in \ac{htc}.
In this experiment, we will evaluate the component responsible for retrieving a subgraph. 
For a comparative assessment, we remove the \ac{rag} system that previously was responsible for fetching a subgraph from the entire knowledge graph while keeping the other parts of the system unchanged.
Instead, we hand the full knowledge graph to the \ac{llm}. This adjusted setup is referred to as Full-KG.

The corresponding performance results are presented in Table~\ref{tab:ab_full}. 
Our KG-HTC has substantially better results in F1-macro than Full-KG, except in the classification of the first level.
The reason is that the total number of labels in the first-level classification is smaller.
The complete knowledge graph hence enables the \ac{llm} to grasp better the semantic relationships among labels, which can consequently aid Full-KG in achieving improved classification performance.
However, as the label space expands, the subgraph retrieval allows \acp{llm} to extract essential information, helping to alleviate performance declines associated with long text inputs.

\input{tables/ablation_qwen}

\paragraph{Performance in open-source LLMs.}~In this experiment, we changed the LLM from \texttt{GPT-3.5-turbo} to \texttt{Qwen2.5-8b}. Table~\ref{tab:ab_openllm} demonstrates that our KG-HTC increases the performance of \texttt{Qwen2.5-8b} except for the first-level classification in the Amazon dataset. This enhancement shows that our KG-HTC has generalization on open-source \ac{llm}.

%% file: tables/error.tex
\begin{table}[t]
\renewcommand{\arraystretch}{1.5}
\caption{Error Analysis. The Hit@K of the \ac{rag} system decreases at the second and third levels across all misclassified samples. }
\centering
\vspace{10pt}
\resizebox{\linewidth}{!}{
\begin{tabular}{cccccc}
\toprule  
\multirow{2}{*}{\textbf{Dataset}} & \textbf{WoS}   & \multicolumn{2}{c}{\textbf{Dbpedia}} &   \multicolumn{2}{c}{\textbf{Amazon}}    \\
\cmidrule(lr){2-2}
\cmidrule(lr){3-4}
\cmidrule(lr){5-6}
               & \textbf{Level 2} & \textbf{Level 2} & \textbf{Level 3} & \textbf{Level 2} & \textbf{Level 3} \\
\toprule 
\textbf{Hit@K} &    0.731     &    0.760     &   0.649      &     0.498    &    0.392     \\
\toprule  
\end{tabular}
}
\label{tab:error}
\vspace{5pt}
\end{table}

%% file: tables/ablation_rag.tex
\begin{table}[t]
\renewcommand{\arraystretch}{1.5}
\caption{Ablation study on the effectiveness of the subgraph retrieval. Full-KG replaces the \ac{rag}-based subgraph retrieval by the retrieval of the entire knowledge graph.}
\centering
\vspace{10pt}
\resizebox{\linewidth}{!}{
\begin{tabular}{ccccccccc}
\toprule  
\multirow{2}{*}{\textbf{Dataset}} & \multicolumn{2}{c}{\textbf{WoS}}   & \multicolumn{3}{c}{\textbf{Dbpedia}} &   \multicolumn{3}{c}{\textbf{Amazon}}    \\
\cmidrule(lr){2-3}
\cmidrule(lr){4-6}
\cmidrule(lr){7-9}
               & \textbf{Level 1} & \textbf{Level 2} & \textbf{Level 1} & \textbf{Level 2} & \textbf{Level 3} & \textbf{Level 1} & \textbf{Level 2} & \textbf{Level 3}\\
\toprule 
Full-KG         &   \textbf{0.762} &   0.616  &  \textbf{0.913}  &    0.670   &  0.884 & \textbf{0.926} & 0.633 & 0.431   \\
\textbf{KG-HTC} &   0.749 &   \textbf{0.651}  &  0.886  &    \textbf{0.811}   &   \textbf{0.902}   &   0.901  &  \textbf{0.651}    &    \textbf{0.462}     \\

\toprule  
\end{tabular}
}
\label{tab:ab_full}
\vspace{5pt}
\end{table}

%% file: tables/ablation_qwen.tex
\begin{table}[t]
\renewcommand{\arraystretch}{1.5}
\caption{Ablation study on the effectiveness of an open-souece LLM \texttt{Qwen2.5-8b}.}
\centering
\vspace{10pt}
\resizebox{\linewidth}{!}{
\begin{tabular}{ccccccccc}
\toprule  
\multirow{2}{*}{\textbf{Dataset}} & \multicolumn{2}{c}{\textbf{WoS}}   & \multicolumn{3}{c}{\textbf{Dbpedia}} &   \multicolumn{3}{c}{\textbf{Amazon}}    \\
\cmidrule(lr){2-3}
\cmidrule(lr){4-6}
\cmidrule(lr){7-9}
               & \textbf{Level 1} & \textbf{Level 2} & \textbf{Level 1} & \textbf{Level 2} & \textbf{Level 3} & \textbf{Level 1} & \textbf{Level 2} & \textbf{Level 3}\\
\toprule 
\textbf{Qwen}         &   0.550 &   0.367  &  0.472  &    0.466   &  0.691 & \textbf{0.783} & 0.326 & 0.230   \\
\textbf{Qwen KG-HTC} &   \textbf{0.715} &   \textbf{0.509}  &  \textbf{0.699}  &     \textbf{0.575} &  \textbf{0.799}    &  0.762   &   \textbf{0.418}   &    \textbf{0.343}     \\
\toprule 
\end{tabular}
}
\label{tab:ab_openllm}
\vspace{5pt}
\end{table}

%% file: chapters/related_work.tex
\section{Related Work}
\label{sec:5}

\paragraph{Hierarchical Text Classification.} The task of \ac{htc} was initially proposed by \citet{sun-2001-hierarchical}, who suggested using Support Vector Machine classifiers as a solution. Subsequently, \citet{kowsari-2017-hdl} explored various deep learning methods, training different deep neural networks for each level of the taxonomy. More recently, \citet{liu-2023-enhancing} introduced the integration of knowledge graphs to enhance \ac{htc}. They employed Graph Neural Networks (GraphSAGE) to encode knowledge graphs and combined graph embeddings with BERT text embeddings for fine-tuning. This approach currently represents the state-of-the-art in supervised settings. However, supervised methods face challenges in industrial applications. The primary issue is the lack of annotated data, as the cost of labeling data across multiple hierarchical levels can be prohibitively high.
As the performance of \ac{llm} continues to improve~\citep{touvron-2023-llama, liu-2024-deepseek, radford-2018-improving, yang-2024-qwen2}, zero-shot inference has become increasingly popular in many fields~\citep{brown-2020-language, hendrycks-2021-measuring, li-2024-autokaggle, zhu-2024-lime}.

\paragraph{Zero-shot \ac{htc}.} Three recent studies align with our strict zero-shot learning setting of \ac{htc}. 
First, \citet{halder-2020-task} proposed the Task-Aware Representation method for zero-shot text classification. 
This approach allows any text classification task to be converted into a universal binary classification problem.
Given a text and a set of labels, the \ac{llm} determines whether a label matches (outputting \texttt{True} or \texttt{False}). However, due to \ac{htc}'s extensive label space, this method requires multiple iterations to complete a single classification. 

Second, \citet{bongiovanni-2023-zero} proposed Zero-shot Semantic Text Classification (Z-STC), which utilized the Upward Score Propagation (USP) method for \ac{htc}. 
This approach leverages pre-trained deep language models to independently encode documents and taxonomy labels into a semantic vector space, such as BERT \cite{devlin-2019-bert} and RoBERTa~\citep{liu-2019-roberta}. 
The prior relevance score of each label is then computed via cosine similarity. 
By incorporating the hierarchical structure of the taxonomy, USP recursively propagates the relevance scores of lower-level labels upward to their parent nodes. 
The key idea is that if a child label is highly relevant to the content to be classified, its parent label should also be considered relevant. 
This hierarchical score propagation effectively integrates local semantic cues into the global taxonomy, improving overall classification performance.

Finally, \citet{paletto-etal-2024-label} proposed the Hierarchical Label Augmentation (HiLA) method for \ac{htc}.
They used pre-trained \ac{llm} to generate additional child labels for the leaf nodes of the existing label hierarchy. 
Since satisfying token constraints by directly inputting the whole hierarchy into an LLM is challenging, they adopted an iterative approach to generate extended sub-labels for each branch to avoid duplication and label overlap.
Moreover, \citet{paletto-etal-2024-label} applied Bongiovanni's Z-STC on the new deeper levels to accomplish the \ac{htc} task.
As such, HiLA represents the state-of-the-art in the strict zero-shot setting.
However, as we demonstrate in Section~\ref{sec:4}, Bongiovanni's \citep{bongiovanni-2023-zero} and Paletto's \cite{paletto-etal-2024-label}'s methods exhibit low classification accuracy for deeper-level labels. As the taxonomy deepens, the label space for the corresponding single-layer classification task becomes larger, and the distributions of labels become more biased.

\paragraph{Retrieval Augment Generation.}~\ac{rag} can dynamically retrieve relevant information from an external corpus or database during in-context learning inference \citep{lewis-2020-retrieval}. 
Graph \ac{rag} extends \ac{rag} to retrieve interconnected entities and relationships through \acp{kg}, enabling richer contextual understanding \citep{edge-2024-local}.
Both \ac{rag} and graph \ac{rag} have demonstrated significant improvements in open-domain question answering \citep{fan-2024-survey, peng-2024-graph}. 
However, there is limited research on leveraging \ac{rag} and graph \ac{rag} for text classification. 

%% file: chapters/conclusion.tex
\section{Conclusion}
\label{sec:6}
Supervised methods for \ac{htc} face challenges in industrial applications.
This paper presents KG-HTC, a zero-shot \ac{htc} method that leverages knowledge graphs and \acp{llm} for \ac{htc}.
Our proposed KG-HTC dynamically retrieves subgraphs that are semantically relevant to the input text and constructs structured prompts, thereby significantly enhancing classification performance under strict zero-shot scenarios. 
Experimental results demonstrate that KG-HTC achieves state-of-the-art performance on three open datasets (Amazon, WoS, and Dbpedia) in the zero-shot setting, with significant improvements observed in the classification of deeper hierarchical labels.

\section{Limitations}
The limitation of this study is the assumption that a complete and correct label taxonomy is available. Consequently, any inaccuracies or errors within the knowledge graph are likely to result in a degradation of our method's performance. Our future research will focus on leveraging large language models to construct robust and accurate knowledge graphs in a more efficient manner. 
And our evaluation datasets may be used in post-training stages of LLMs.

\section*{Ethics Statement}
No ethics issues. \ac{llm} Databases and our evaluation datasets from Kaggle can be used for research purposes.